\documentclass{sig-alternate-modified}

\usepackage{url}
\usepackage{subfigure}
\usepackage{graphicx}
\usepackage{pdfpages}

\begin{document}

\title{Roborobo! a Fast Robot Simulator \\ for Swarm and Collective Robotics}

\numberofauthors{4}
\author{
\alignauthor
	Nicolas Bredeche\\
     \affaddr{UPMC, CNRS}\\
	 \affaddr{Paris, France}\\
      \affaddr{nicolas.bredeche@isir.upmc.fr}
\alignauthor
	Jean-Marc Montanier\\
	 \affaddr{NTNU}\\
	 \affaddr{Trondheim, Norway}\\
      \affaddr{montanier.jeanmarc@gmail.com}
\and
\alignauthor
	Berend Weel\\
	 \affaddr{Vrije Universiteit Amsterdam}\\
	 \affaddr{The Netherlands}\\
      \affaddr{b.weel@vu.nl}
\alignauthor
	Evert Haasdijk\\
	 \affaddr{Vrije Universiteit Amsterdam}\\
	 \affaddr{The Netherlands}\\
      \affaddr{e.haasdijk@vu.nl}
}

\date{April 10th, 2013}

\maketitle
\begin{abstract}

Roborobo! is a multi-platform, highly portable, robot simulator for large-scale collective robotics experiments. Roborobo! is coded in C++, and follows the KISS guideline ("Keep it simple"). Therefore, its external dependency is solely limited to the widely available SDL library for fast 2D Graphics. Roborobo! is based on a Khepera/ePuck model. It is targeted for fast single and multi-robots simulation, and has already been used in more than a dozen published research mainly concerned with evolutionary swarm robotics, including environment-driven self-adaptation and distributed evolutionary optimization, as well as online onboard embodied evolution and embodied morphogenesis.

\end{abstract}



\section{Introduction}

\begin{figure*}
\centering
\begin{tabular}{ccc}
\includegraphics[width=0.3\linewidth]{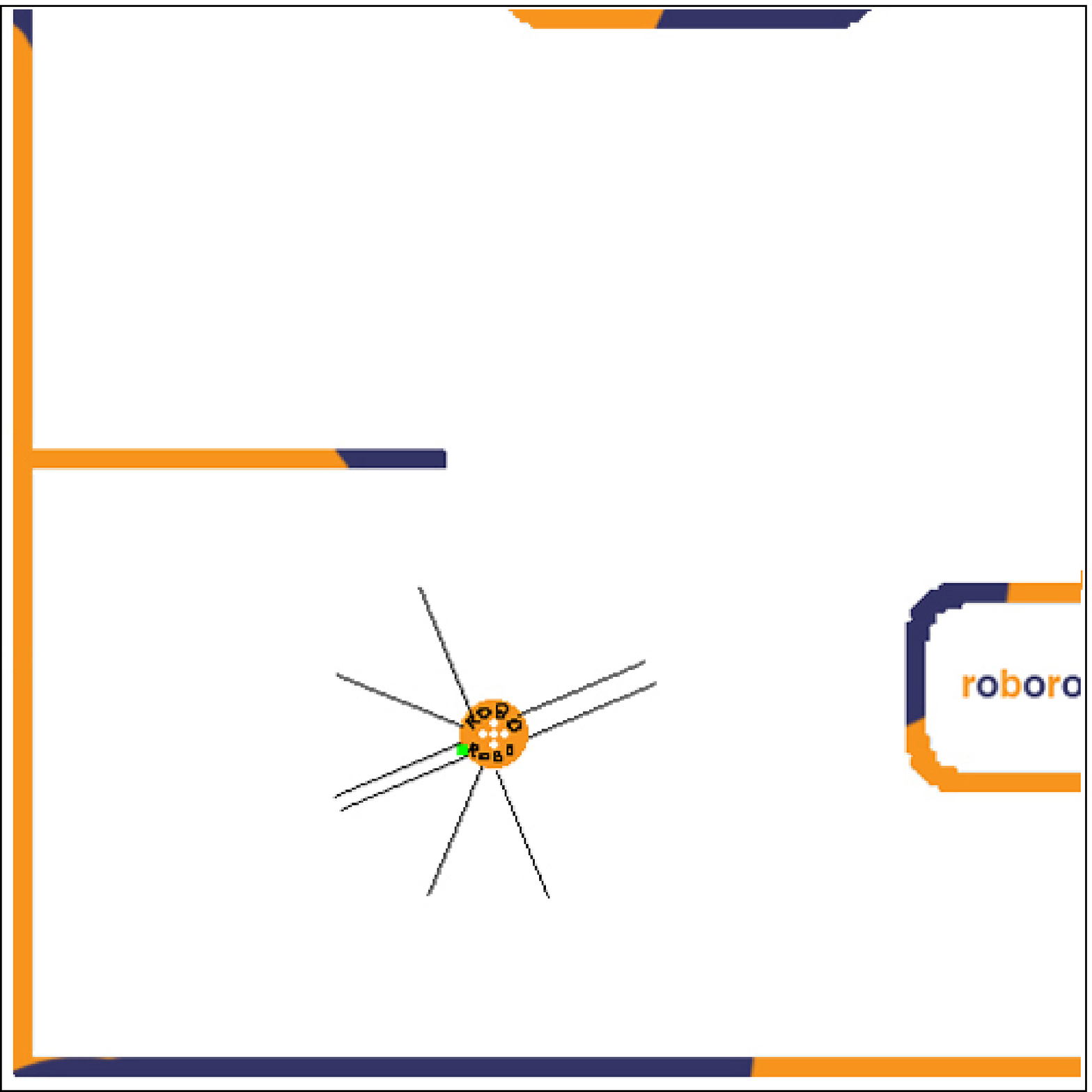} & 
\includegraphics[width=0.3\linewidth]{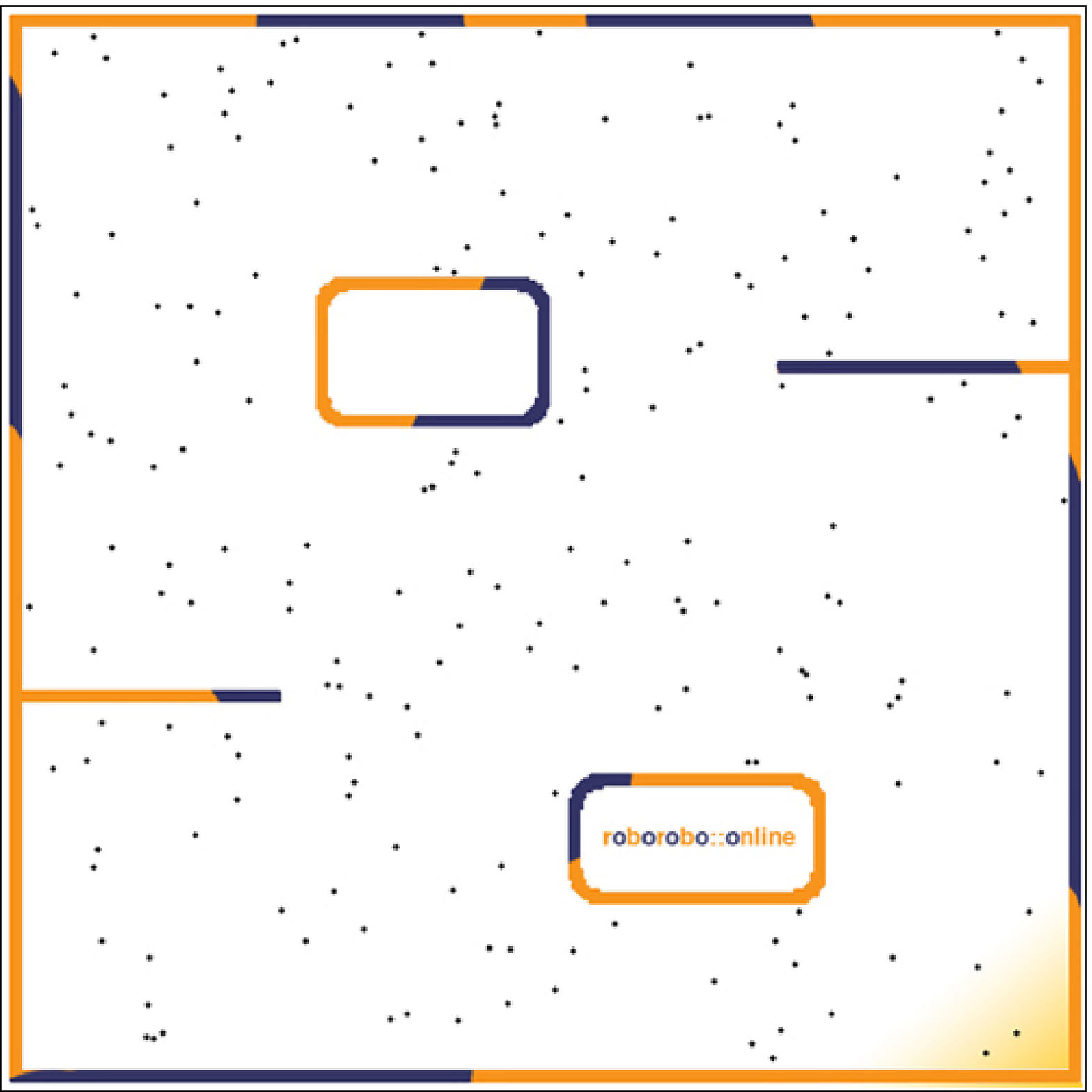} &
\includegraphics[width=0.3\linewidth]{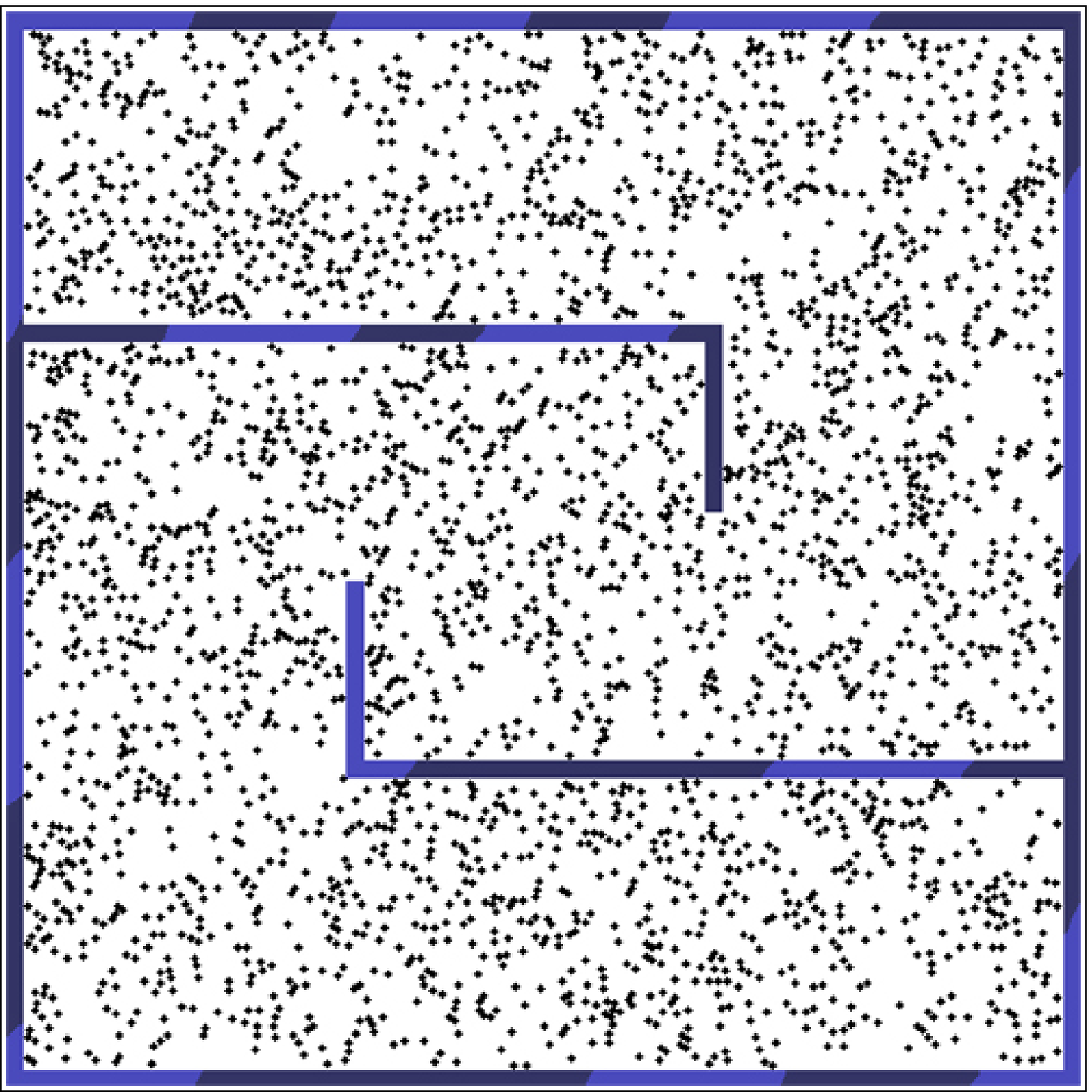} \\
\end{tabular}
\caption{Classic examples with $1$ virtual (e-puck) agent, incl. IR sensor rays (left) ; $100$ virtual agents (center) ; $5000$ virtual agents (right).}
\label{Roborobo-demo}
\end{figure*}

This paper introduces Roborobo!, a light-weight multi-platform simulator for extensive robotics experiment, based on basic robotic hardware setup similar to the famous ePuck~\cite{epuck} or Khepera~\cite{khepera} robotic platforms. Roborobo! is mostly intended at, but not limited to, researchers and practictionners interested in collective adaptive systems and evolutionary robotics, with a particular emphasis on embodied evolution (ie. online evolution) and swarm robotics ($>100$ robots, as illustrated in Figure~\ref{Roborobo-demo}). The underlying idea is to provide a versatile framework for easy development of new ideas and very fast and robust implementation for extensive experiments. Roborobo! is available under the new BSD Licence~\cite{newBSD} at \url{https://code.google.com/p/Roborobo/}.

With respect to other robotic simulators, Roborobo! takes an intermediate approach to model a robotic setup in order to combine (pseudo-)realistic modelling with fast-paced simulation. As such, it stands inbetween realistic, but slow, robotic simulation framework (such as Player/Stage~\cite{playerstage}, Webots~\cite{webots}, V-Rep~\cite{v-rep}, Gazebo~\cite{gazebo} or Microsoft Robotic Developper Studio~\cite{mrds}), and unrealistic, but easy to use, agent-based simulation tools such as Netlogo~\cite{netlogo} and MASON~\cite{mason}. It also differs from easy-access and fast robotic agent simulators such as Breve~\cite{breve} and Simbad~\cite{simbad}, by focusing solely on swarm and aggregate of robotic units, focusing on large-scale population of robots\footnote{e.g. Roborobo! is able to run with up to $6000$ robots on a 4GB laptop computer.} in 2-dimensional worlds rather than more complex 3-dimensional models.

Roborobo! is written in $C++$ with the multi-platform SDL graphics library~\cite{sdl} as unique dependency, enabling easy and fast deployment. It has been tested on a large set of platforms (PC, Mac, OpenPandora~\cite{openpandora}, Raspberry~Pi~\cite{rpi}) and operating systems (Linux, MacOS~X, MS Windows). It has also been deployed on large clusters, such as the french Grid5000~\cite{grid5000} national clusters. Roborobo! has been initiated in 2009 and is used on a daily basis in several universities for both research and education, including Universite Paris-Sud and Universite Pierre et Marie Curie (France), Vrije Universiteit Amsterdam (NL), and NTNU (Norway). 

Roborobo! was originaly developped by Nicolas Bredeche in $2009$ in the context of the european Symbrion Integrated Project~\cite{symbrion}, and has been continuously extended since then. It has been extensively used in various contexts, mostly concerned with evolutionary robotics and swarm robotics, including embodied evolution~\cite{Huijsman2011An-On-line-On-b,Garcia-Sanchez2012Testing-Diversi}, environment-driven evolutionary adaptation~\cite{montanier09evoderob,Bredeche10mEDEA,Montanier11ecal,Montanier11gecco,bredeche12mcmds,Bredeche12ALife,Noskov2013MONEE:-Multi-Ob} and self-assembly~\cite{weel2012organisms,Berend-Weel2013Body-Building}.


\section*{Acknowledgments}
{\small The authors wish to thank all people who contributed to this project, including Leo Cazenille. This work was made possible by the European Union FET Proactive Initiative: Pervasive Adaptation funding the Symbrion project under grant agreement 216342. Roborobo! was extensively tested in a cluster environment using the Grid'5000 experimental testbed, being developed under the INRIA ALADDIN development action with support from CNRS, RENATER and several Universities as well as other funding bodies (see https://www.grid5000.fr).}

\bibliographystyle{abbrv}
{\small
\bibliography{Roborobo-bib}}

\end{document}